\documentclass{article}

\usepackage{arxiv}

\usepackage[utf8]{inputenc} 
\usepackage[T1]{fontenc}    
\usepackage{hyperref}       
\usepackage{url}            
\usepackage{booktabs}       
\usepackage{amsfonts}       
\usepackage{nicefrac}       
\usepackage{microtype}      
\usepackage{lipsum}		
\usepackage{graphicx}
\usepackage{natbib}
\usepackage{doi}
\usepackage{adjustbox}
\usepackage{caption}
\usepackage[table,xcdraw]{xcolor}
\usepackage{booktabs}
\hypersetup{
    colorlinks=false,
    pdfborder={0 0 0},
    pdfborderstyle={/S/U/W 0},
}

\title{Manipulation Risks in Explainable AI: The Implications of the Disagreement Problem}

\author{ \href{https://orcid.org/0000-0003-3784-826X}{\includegraphics[scale=0.06]{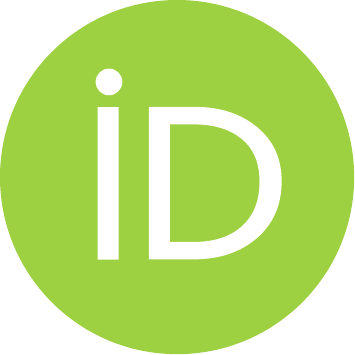}\hspace{1mm}Sofie~Goethals} \\
	Department of Engineering Management\\
	University of Antwerp\\
	Antwerp, Belgium \\
	\texttt{sofie.goethals@uantwerpen.be} \\
	\And
        David Martens\\
	Department of Engineering Management\\
	University of Antwerp\\
	Antwerp, Belgium \\
	\And
        Theodoros Evgeniou\\
        Decision Sciences and Technology Management Department\\
        INSEAD\\
        Fontainebleau, France\\
}

\renewcommand{\shorttitle}{{Manipulation Risks in XAI}}

\hypersetup{
pdftitle={Manipulation Risks in XAI},
pdfauthor={Sofie~Goethals, David~Martens, Theodoros~Evgeniou},
pdfkeywords={Explainable AI (XAI), Manipulation, Responsible AI},
}

\begin{document}
\maketitle

\begin{abstract}
	Artificial Intelligence (AI) systems are increasingly used in high-stakes domains of our life, increasing the need to explain these decisions and to make sure that they are aligned with how we want the decision to be made. 
The field of Explainable AI (XAI) has emerged in response. However, it faces a significant challenge known as the disagreement problem, where multiple explanations are possible for the same AI decision or prediction. While the existence of the disagreement problem is acknowledged, the potential implications associated with this problem have not yet been widely studied. First, we provide an overview of the different strategies explanation providers could deploy to adapt the returned explanation to their benefit. We make a distinction between strategies that \emph{attack} the machine learning model or underlying data to influence the explanations, and strategies that \emph{leverage} the explanation phase directly. Next, we analyse several objectives and concrete scenarios the providers could have to engage in this behavior, and the potential dangerous consequences this manipulative behavior could have on society.   We emphasize that it is crucial to investigate this issue now, before these methods are widely implemented, and propose some mitigation strategies.
\end{abstract}


\section{Introduction}
Artificial Intelligence (AI) is used in more and more high-stakes domains of our life such as justice~\citep{berk2012criminal}, healthcare~\citep{callahan2017machine}, and finance~\citep{lessmann2015benchmarking}, increasing the need to explain these decisions and to make sure that they are aligned with how we want the decision to be made. However, the complexity of many AI systems makes them challenging to comprehend, posing a significant barrier to their implementation and oversight~\citep{arrieta2020explainable,samek2019explainable}.  Legislative initiatives, including the EU General Data Protection Regulation (GDPR), have recognized the `right for explanation' for individuals affected by algorithmic-decision making, emphasizing the legal necessity of explainability~\citep{goodman2017european}.
In response, the field of Explainable Artificial Intelligence (XAI) has emerged, aimed at developing methods for explaining the decision-making processes of AI models~\citep{adadi2018peeking,holzinger2022explainable,xu2019explainable}.

Nevertheless, the landscape of post-hoc explanations is diverse, and each method can yield a different explanation. Furthermore, even within a single explanation method, multiple explanations can be generated for the same instance or decision. This phenomenon, known as the \emph{disagreement problem}, has been studied in literature~\citep{brughmans2023disagreement,krishna2022disagreement,neely2021order,roy2022don}. While the existence of the disagreement problem is acknowledged, the potential implications of this problem have not yet been extensively explored. \citet{barocas2020hidden} already mention that the power to choose which explanation to return, leaves the providers with significant room to promote their own welfare. \citet{aivodji2019fairwashing} discuss the possibility of fairwashing, where discriminatory practices can be hidden by selecting the right explanations, while \citet{bordt2022post} argue that post-hoc explanations fail to achieve their purpose in adversarial contexts. However, an overview of potential misuses by the explanation provider is still missing from the literature, and we believe it is imperative to study the implications now, before explainability methods are implemented on a wide scale.
The main contributions of this paper are:
\begin{itemize}
    \item Providing a comprehensive framework that outlines the different strategies that could be employed by malicious entities to manipulate the  explanations.
    \item An overview of the different objectives these actors could have to engage in this behavior, and the potential implications.
\end{itemize}
This paper is structured as follows: We introduce the field of Explainable AI and the disagreement problem in Sections~\ref{sec:xai} and \ref{sec:disagreement}. In Section~\ref{sec:strategies}, we explore various strategies that providers could employ to manipulate the explanations according to their preferences. Additionally, in Section~\ref{sec:objectives}, we present specific objectives and scenarios that may drive providers to engage in such behavior. 
Finally, in Section~\ref{sec:discussion}, we offer discussion and potential solutions to address this.


\section{Explainable AI} \label{sec:xai}
Within the field of Artificial Intelligence, providing insights into the decision-making process is crucial for various reasons.  First, it establishes trust and compliance with stakeholders, as they can understand and validate the reasoning behind the model's output. Secondly, it enables improved domain insights, allowing practitioners and users to gain a deeper understanding of the problem space and uncover valuable knowledge. Lastly, insights derived from explanations can lead to model improvement, aiding in the optimization of AI systems~\citep{molnar2020interpretable,xu2019explainable}.

To reach these goals, various methods to achieve comprehensibility in AI models have been proposed.
In general, there are two main approaches commonly used:
 inherently transparent models and post-hoc explanations. Inherently transparent models, such as small decision trees, are comprehensible by nature due to their simple structure, without the need for additional explanations~\citep{molnar2020interpretable}.
 However, in many real-world scenarios, data is becoming increasingly complex and black-box models are used due to their superior predictive performance~\citep{goethals2022non}. These models lack inherent interpretability, and post-hoc explanations are used to provide insights into their decision-making process. This field of research is commonly known as Explainable Artificial Intelligence (XAI).
 
 Within XAI, a distinction can be made between global and local explanations. Global explanations aim to provide an understanding of the model's logic as a whole, allowing users to follow the reasoning that leads to every possible outcome. Techniques such as rule extraction~\citep{martens2007comprehensible} and Partial Dependence plots~\citep{friedman2001greedy} fall under this category. On the other hand, local post-hoc explanations focus on explaining the logic behind a specific prediction or decision made by the model. Methods like SHAP (SHapley Additive exPlanations)~\citep{lundberg2017unified} and LIME (Local Interpretable Model-agnostic Explanations)~\citep{ribeiro2016should} are examples of post-hoc explanation that measure the impact of each feature for a given prediction score (feature importance methods). Another local technique, known as counterfactual explanations, describes a combination of feature changes required to alter the predicted class~\citep{martens2014explaining,wachter2017counterfactual,guidotti2022counterfactual}. 
 While this paper predominantly uses counterfactual explanations as an example, the findings and discussion presented are applicable to other post-hoc explanation techniques as well. At the moment, we do not see manipulation issues for inherently transparent models but this would be an interesting avenue for future research~\citep{bordt2022post}.

 In line with \citet{greene2023monetizing}, we define an explanation recipient as a person who requests an explanation for an automated decision, and an explanation provider as the entity who provides the algorithmic explanations to the recipient. For example, in the domain of finance, the explanation provider could be a bank, and the explanation recipient a loan applicant; while in the domain of employment the explanation recipient would be the job applicant, and the explanation provider the hiring agency~\citep{greene2023monetizing}. Not all scenarios described in Section~\ref{sec:objectives} assume that there is one actual recipient; the explanation provider can also provide explanations of the model to the public proactively or to comply with regulatory requirements.

\section{The Disagreement Problem} \label{sec:disagreement}

\begin{table}[]
 \centering
\begin{adjustbox}{max totalsize={1\textwidth}{0.65\textheight},center}\begin{tabular}{@{}lcccccccccccccc@{}}
\toprule
             & Sex                       & Age                         & \begin{tabular}[c]{@{}c@{}}Residence\\ time\end{tabular} & \begin{tabular}[c]{@{}c@{}}Home\\ status\end{tabular} & Occupation                & \begin{tabular}[c]{@{}c@{}}Job\\ status\end{tabular} & \begin{tabular}[c]{@{}c@{}}Employment\\ time\end{tabular} & \begin{tabular}[c]{@{}c@{}}Other\\ investments\end{tabular} & \begin{tabular}[c]{@{}c@{}}Bank\\ account\end{tabular} & \begin{tabular}[c]{@{}c@{}}Time\\ at bank\end{tabular} & Liability                 & \begin{tabular}[c]{@{}c@{}}Account\\ reference\end{tabular} & \begin{tabular}[c]{@{}c@{}}Housing\\ expense\end{tabular} & \begin{tabular}[c]{@{}c@{}}Savings\\ account\end{tabular} \\ \midrule

Instance     & 2                         & 16                          & 22                                                       & 1                                                     & 2                         & 6                                                    & 7                                                            & 0                                                           & 0                                                      & 0                                                      & 0                         & 1                                                           & 1                                                         & 125                                                       \\ \midrule
CBR          & 2                         & 16                          & \cellcolor[HTML]{C0C0C0} 0.25                                                     & 1                                                     & 2                         & 6                                                    & 7                                                            & 0                                                           & \cellcolor[HTML]{C0C0C0}1                              & 0                                                      & 0                         & 1                                                           & 1                                                         & 125                                                       \\
DiCE         & 2                         & 16                          & 22                                                       & 1                                                     & 2                         & 6                                                    & 7                                                            & \cellcolor[HTML]{C0C0C0}24                                  & 0                                                      & 0                                                      & 0                         & 1                                                           & 1                                                         & 125                                                       \\
GeCo         &                           &                             &                                                          &                                                       &                           &                                                      &                                                              &                                                             &                                                        &                                                        &                           &                                                             &                                                           &                                                           \\
NICE(none)   & 2                         & \cellcolor[HTML]{C0C0C0}34  & \cellcolor[HTML]{C0C0C0}0                                & \cellcolor[HTML]{C0C0C0}3                             & \cellcolor[HTML]{C0C0C0}3 & \cellcolor[HTML]{C0C0C0}10                           & \cellcolor[HTML]{C0C0C0}1                                    & 0                                                           & 0                                                      & 0                                                      & 0                         & 1                                                           & \cellcolor[HTML]{C0C0C0}2                                 & \cellcolor[HTML]{C0C0C0}136                               \\
NICE(plaus)  & 2                         & \cellcolor[HTML]{C0C0C0}34  & \cellcolor[HTML]{C0C0C0}0                                & \cellcolor[HTML]{C0C0C0}3                             & \cellcolor[HTML]{C0C0C0}3 & 6                                                    & \cellcolor[HTML]{C0C0C0}1                                    & 0                                                           & 0                                                      & 0                                                      & 0                         & 1                                                           & \cellcolor[HTML]{C0C0C0}2                                 & \cellcolor[HTML]{C0C0C0}136                               \\
NICE(prox)   & 2                         & \cellcolor[HTML]{C0C0C0}34  & \cellcolor[HTML]{C0C0C0}0                                & 1                                                     & 2                         & \cellcolor[HTML]{C0C0C0}10                           & \cellcolor[HTML]{C0C0C0}1                                    & 0                                                           & 0                                                      & 0                                                      & 0                         & 1                                                           & 1                                                         & \cellcolor[HTML]{C0C0C0}136                               \\
NICE(sparse) & 2                         & 16                          & \cellcolor[HTML]{C0C0C0}0                                & 1                                                     & 2                         & \cellcolor[HTML]{C0C0C0}10                           & \cellcolor[HTML]{C0C0C0}1                                    & 0                                                           & 0                                                      & 0                                                      & 0                         & 1                                                           & 1                                                         & \cellcolor[HTML]{C0C0C0}136                               \\
SEDC         & 2                         & 16                          & 22                                                       & 1                                                     & 2                         & 6                                                    & 7                                                            & 0                                                           & \cellcolor[HTML]{C0C0C0}1                              & 0                                                      & 0                         & 1                                                           & 1                                                         & 125                                                       \\
WIT          & \cellcolor[HTML]{C0C0C0}1 & \cellcolor[HTML]{C0C0C0}278 & \cellcolor[HTML]{C0C0C0}8                                & \cellcolor[HTML]{C0C0C0}2                             & \cellcolor[HTML]{C0C0C0}1 & \cellcolor[HTML]{C0C0C0}5                            & \cellcolor[HTML]{C0C0C0}1                                    & \cellcolor[HTML]{C0C0C0}6.5                                 & \cellcolor[HTML]{C0C0C0}1                              & \cellcolor[HTML]{C0C0C0}1                              & \cellcolor[HTML]{C0C0C0}6 & \cellcolor[HTML]{C0C0C0}0                                   & \cellcolor[HTML]{C0C0C0}0                                 & \cellcolor[HTML]{C0C0C0}102                               \\ \bottomrule
\vspace{1mm}
\end{tabular}
\end{adjustbox}
\caption{Illustration of the disagreement problem for an instance of the Australian Credit dataset.}
\label{tab:exampleinstance}
\end{table}
A known issue within Explainable AI is that the results of different explanation techniques do not always agree with each other. Even one explanation technique can generate many different explanations for one instance, which is known as the disagreement problem~\citep{krishna2022disagreement,neely2021order,roy2022don}. 
One of the reasons behind the disagreement problem is that a `true internal reason' why the machine learning model comes to a certain decision, generally does not exist~\citep{bordt2022post}. For example, for feature importance methods such as SHAP and LIME, there is no mathematically unique way to determine the importance of each feature to the decision of a black-box function~\citep{bordt2022post,sundararajan2020many}. As a consequence, all feature importance methods rely on their own assumptions to approximate this~\citep{bordt2022post,sundararajan2020many}. For counterfactual explanations, this issue also exists as the optimization problem to create the explanations can be set up in a different way in every implementation. Even a single counterfactual explanation method could lead to a large number of explanations, as the choice of parameters (such as the distance metric) has an influence on the explanations that are returned first~\citep{goethals2023precof}. The diversity of multiple counterfactual explanations, generated by the same counterfactual algorithm is also known as the Rashomon effect~\citep{molnar2020interpretable}.\footnote{The Rashomon effect means that an event can be explained by multiple causes, and is named  after a Japanese movie that tells multiple (contradictory) stories about the death of a samurai~\citep{molnar2020interpretable}.}

Other authors already showed the level of disagreement between different post-hoc explanation techniques: \citet{roy2022don} show disagreement between LIME and SHAP explanations, \citet{brughmans2023disagreement} illustrate this for different counterfactual explanation algorithms, and \citet{bordt2022post} demonstrate the disagreement between SHAP, LIME, and counterfactual explanations. 
We illustrate the disagreement problem for one specific instance with an example in Table~\ref{tab:exampleinstance}, in line with \citet{brughmans2023disagreement}. This table demonstrates the disagreement problem between different counterfactual explanation algorithms for one instance from the Australian credit dataset, where the target variable indicates whether a person should be granted a loan or not. The depicted instance was not awarded credit and asks for a counterfactual explanation to know which features to change to receive a positive credit decision. Table~\ref{tab:exampleinstance} shows the explanations returned by 10 different counterfactual algorithms, which vary widely. \footnote{The counterfactual algorithm GeCo was not able to find a counterfactual explanation for the given instance.}
This example illustrates that every feature can be included in the explanation by switching between explanation algorithms. \citet{brughmans2023disagreement} verify this for multiple datasets and classifiers, and establish the feasibility of both including and excluding specific features across different scenarios. Note that the potential for manipulation of explanations extends beyond switching between different counterfactual explanation algorithms. In Section~\ref{sec:strategies}, alternative strategies that can be employed for manipulation are explored.
Currently, a consensus on how to resolve this ambiguity has not yet been reached.
 Research indicates that most developers rely on arbitrary heuristics, such as personal preferences, to choose the final explanation~\citep{krishna2022disagreement}.

This plurality is not necessarily a bad thing. \citet{bordt2022post} distinguish between a cooperative and an adversarial context. In cooperative contexts, where stakeholders have the same goal, this plurality can be beneficial as it is expected that the explanation provider will choose the explanation that is in both parties' best interest. For example, when data scientists are debugging a model for their own company, this plurality of explanations can be useful.  However, in adversarial contexts, the interests of the explanation provider and the data subject are not necessarily aligned, and the explanation providers will be incentivized to choose the explanation that best fits their own interests. An example of such an adversarial context is a loan application where the customer was denied the loan and wants to flag the decision as being discriminatory~\citep{bordt2022post}. In this case, the bank might want to conceal this discriminatory practice by returning a different explanation. This phenomenon is known as \emph{fairwashing}, and has received significant attention~\citep{aivodji2019fairwashing}. While fairwashing is the most extensively studied objective, we will explore additional scenarios for misuse in adversarial contexts  in Section~\ref{sec:objectives}.
However, even in adversarial contexts, this plurality can be used in a positive way. For example, \citet{bove2023investigating} do mention that in settings such as loan applications, the plurality of explanations can benefit the user if they are provided with multiple explanations. 

\section{Manipulation Strategies: How can explanation providers exploit the disagreement problem?} \label{sec:strategies}
\begin{figure}[h]
    \centering
    \includegraphics[width=\textwidth]{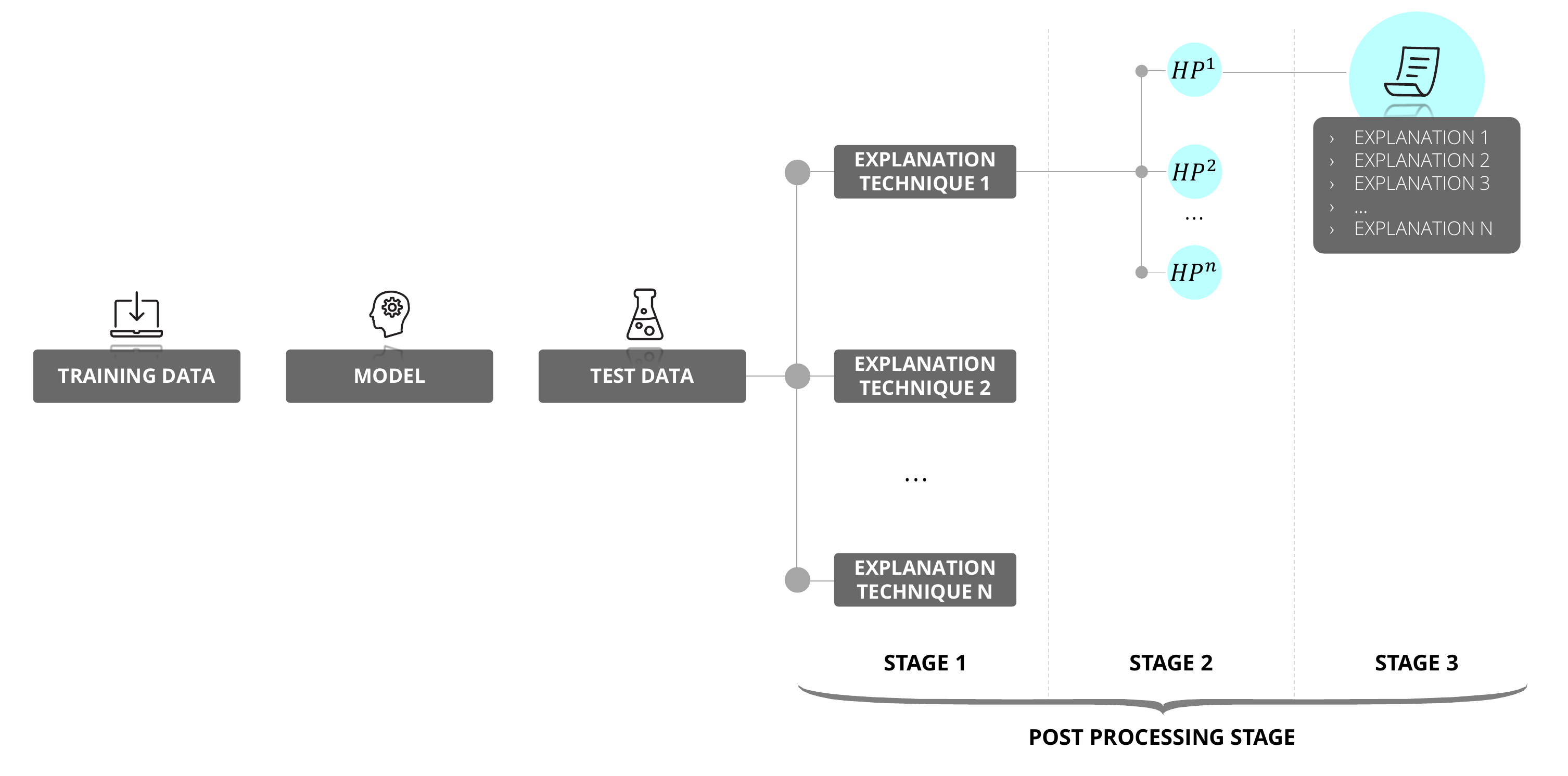}
    \caption{Strategies the explanation providers could deploy to manipulate the explanations}
    \label{fig:strategies}
\end{figure}

The manipulation of explanations by explanation providers is not limited to the mentioned example of switching between explanation algorithms, but can occur at various stages throughout the pipeline, as depicted in Figure~\ref{fig:strategies}. We specifically focus on the manipulation that takes place in the post-processing stage, where the explanations are generated, as we imagine that the explanation provider may not always possess the authority to modify the machine learning model or underlying data (the explanation provider is not necessarily the same entity as the model owner). Nevertheless, it is important to note that manipulations directly to the data or model are still feasible, and we discuss some relevant literature exploring this below.


Manipulating the training data to result in different explanations, is related to the area of \emph{data poisoning attacks}. Data poisoning attacks usually involve injecting manipulated data into the training set to compromise the performance of the machine learning model, and while the main focus in literature is on model behavior, its goal might also be manipulating the explanations. \citet{baniecki2023fooling} illustrate that it is possible to attack Partial Dependence plots by poisoning the training data. \citet{bordt2022post} highlight the important role of the reference dataset, and show how changing this set influences the resulting SHAP explanations. 
With regard to changing the model, \citet{slack2020fooling} demonstrate the possibility of modifying biased classifiers in such a way that they continue to yield biased predictions, while the explanations generated by LIME and SHAP will appear harmless. Other authors show the possibility of fine-tuning a neural network to conceal discrimination in the model explanations~\citep{dimanov2020you,heo2019fooling}.
Finally, in the domain of images, \citet{dombrowski2019explanations} present evidence showcasing the manipulation of explanations through the application of nearly imperceptible perturbations to visual inputs. In this case, the test data, for which the prediction needs to be explained, is altered. These perturbations would not change the output of the machine learning model, but could result in drastic changes in the explanation map. \footnote{One could argue that altering the test data in an imperceptible way will be mostly applicable to image data, as in tabular data these changes may be more noticeable.}
Additionally, \citet{slack2021counterfactual} focus on modifying both the model and the test data, such that slight perturbations to the input data can lead to more cost-effective recourse for specific subgroups, while giving the impression of fairness to auditors.

As mentioned, we focus on strategies to alter the explanation in the post-processing stage, without making any alterations to the used data or the underlying machine learning model. 
We foresee three main strategies the providers could deploy in this stage:
 \begin{enumerate}
     \item \textbf{Change the explanation technique}\\ 
     Many different post-hoc explanation techniques exist, both local and global, as outlined in Section~\ref{sec:xai}. Consequently, a first evident strategy entails switching to a different explanation technique. For example, when the surrogate model reveals patterns the explanation provider wants to conceal, he might switch to using Partial Dependence plots as an alternative if these patterns do not manifest clearly in those plots.
     However, on a local level, using different explanation techniques between instances may attract greater attention than the strategies described below, as the output could have a significantly different format (e.g., feature importance plot versus a counterfactual explanation).
     \item \textbf{Change the parameters or used implementation of an explanation technique\\} Even within  a single explanation algorithm, significant leeway exists for manipulating the explanations, contingent upon the selected parameter configurations. 
     For example, LIME explanations depend on the number of perturbed instances and the bandwidth~\citep{bordt2022post,garreau2020explaining}, while for Shapley values, there is a multitude of ways to implement them and each operationalization yields significantly different results~\citep{sundararajan2020many}. Global methods, such as surrogate modeling, are heavily influenced by the choice of architectural design (e.g., linear models, decision trees, etc.) and the complexity of the surrogate model. In the case of counterfactual explanations, as shown in Table~\ref{tab:exampleinstance}, the used implementation exerts a substantial influence on the returned explanations, with the number of potential implementations proliferating at a rapid pace. Additionally, even within one counterfactual algorithm, there often exist many modifiable parameters that influence the results.
    \item \textbf{Exploit the non-deterministic component of some explanation algorithms\\} Some explanation algorithms such as DICE~\citep{mothilal2020explaining} inherently provide multiple possible explanations for one instance. In such cases, the explanation provider can simply select an explanation from the available options without requiring any modifications. Furthermore, certain explanation algorithms are not designed in a deterministic way and may return different explanations across runs.  For example, when using LIME, the randomness introduced during the sampling and perturbation process can lead to variations in the generated explanations for each execution~\citep{lee2019developing,zhang2019should}. Additionally, \citet{de2021framework} show that multiple counterfactual algorithms do not generate consistent results over multiple runs, when the same model, input data and parameters are used.
    In this scenario, the explanation providers can repeatedly execute the explanation algorithm until an explanation that aligns with their preferences is returned. 
 \end{enumerate}

 In the scenario we describe, we assume explanation providers deliberately choose the explanation out of all the possible explanations that best aligns with their interests. The returned explanation will still be technically correct, it will just not necessarily be the explanation that will be in the best interest of the user. It is important to note that we are not referring to situations where explanations chosen by the explanation provider are not in the best interest of the user `by accident' due to differences in knowledge background or a lack of awareness of the user's preferences~\citep{bove2023investigating,gilpin2022explanation}. 
 Instead, we are concerned with cases where the explanation provider knowingly opts for an explanation that serves their own agenda, despite knowing that it may not be the optimal explanation for the end user. Note that in described strategies, 
 the providers maintain a partial ethical stance by delivering explanations that retain technical correctness. However, providers have the potential to further exploit the situation by offering spam explanations, containing superfluous features~\citep{greene2023monetizing}, or by deliberately presenting entirely false explanations that are fabricated.
 The complexity of the pipeline depicted in Figure~\ref{fig:strategies} demonstrates the extensive potential for manipulation and, consequently, the fragility of explanations.

\section{Manipulation Objectives: Why would explanation providers want to exploit the disagreement problem?} \label{sec:objectives}

\begin{figure} [ht]
    \centering
    \includegraphics[width=\textwidth]{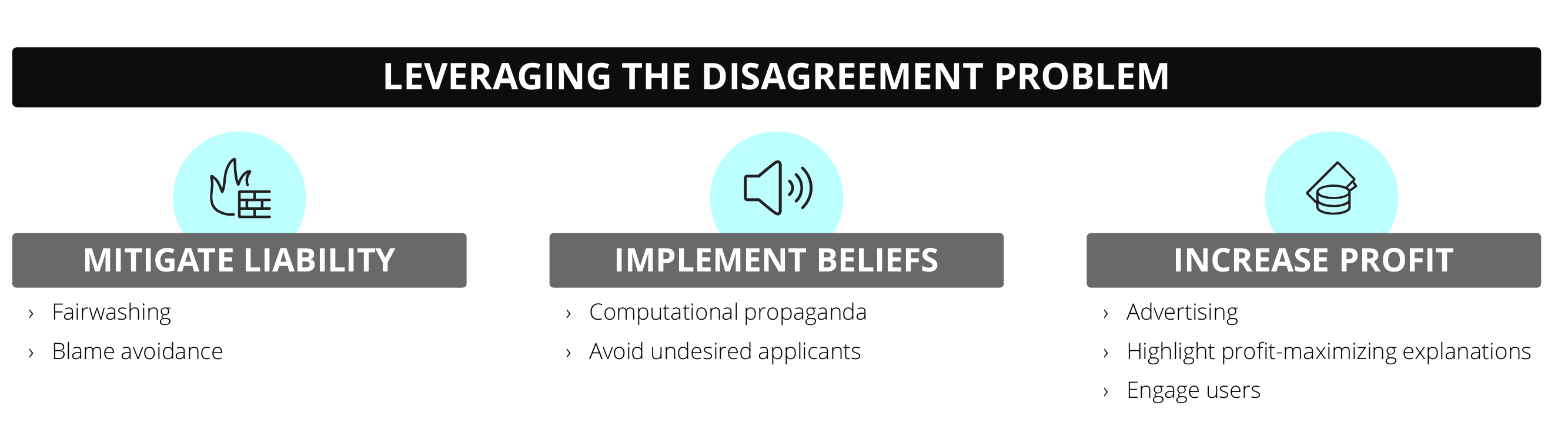}
    \caption{Main objectives to leverage the disagreement problem}
    \label{fig:incentives}
\end{figure}

 Which objectives could the providers have to engage in this behavior? We outline them in Figure~\ref{fig:incentives}, and discuss various scenarios for each objective in the subsections below. At the moment, we see mitigating liability, implementing their beliefs and maximizing their profits as the main objectives. This list may not be exhaustive yet as technology is constantly evolving and new objectives may emerge. 
 
\subsection{Mitigate liability}
The model could be unethical or suboptimal in several ways and model explanations could reveal this. Explanation providers could manipulate the explanations to avoid these issues coming to light.

\subsubsection{Fairwashing} 
The first, and most studied, reason for explanation providers to engage in this behavior, is \emph{fairwashing}~\citep{aivodji2019fairwashing,aivodji2021characterizing,shahin2022washing}. Fairwashing is defined as `\emph{promoting the false perception that a machine learning model used by the company is fair while this might not be so}'~\citep{aivodji2019fairwashing}. In a fairwashing attack, the explanation provider will manipulate the explanations to under report the unfairness of the machine learning model. This has a significant impact on the individuals that received a negative decision based on unfair grounds, as this will deprive them of the possibility to contest it~\citep{aivodji2021characterizing}. The relative easiness with which fairwashing can be executed has already been shown in the literature.~\citep{aivodji2021characterizing,shahin2022washing}. 
Imagine a bank that decides it prefers people from a certain demographic group, and predominantly gives out loans to this group (without a justified reason to do so). It could easily mask this behavior by choosing a different explanation. For example, instead of returning the explanation `\emph{If you would have belonged to a different demographic group, you would have received the loan}', it could return as explanation `\emph{If your income would be double as high, you would have received the loan}', even if the latter explanation is less plausible. Some counterfactual algorithms such as DICE~\citep{mothilal2020explaining} even have as an input parameter the features that can be part of the explanation, so if sensitive features such as demographic attributes are removed from this list, counterfactual explanations will never flag discrimination. We use counterfactual explanations as illustration here, but this objective extends to other explanation techniques as well. All the mentioned techniques in Section~\ref{sec:xai} have the potential to reveal bias within a model (for example a feature importance ranking where the sensitive attribute has a very high score). 
This misleading practice undermines the core principles of algorithmic fairness and hampers efforts towards achieving equitable and just outcomes.

\subsubsection{Blame avoidance}
Explanation providers can also take advantage of the plurality of explanations to shift blame or evade responsibility for controversial or erroneous decisions made by Artificial Intelligence (AI) systems. 
\citet{nissenbaum1996accountability} already mention that placing accountability in a computerized system can be a very obscure process due to the `\emph{problem of many hands}' (many actors and factors contribute to the process, and is not clear which factor ultimately led to the decision). This issue is reflected in the explanations, where different explanations can point to different actors or circumstances.
For example, in the case of autonomous vehicles, AI systems make critical decisions that impact passenger safety. Malicious model owners, such as manufacturers or operators, may downplay system failures or accidents caused by their vehicles. They could selectively present an explanation that attributes the fault to external factors or human error, and as such divert attention from potential design flaws or inadequate safety measures.
Similarly, in the field of healthcare, this exploitative behavior can manifest when mistakes by surgeons or flaws in operating machines are concealed to avoid accountability. Explanation providers, which could include medical professionals, institutions, or even the manufacturers of medical devices, may withhold or manipulate explanations to protect their reputations or evade legal consequences. Such practices can have severe consequences, as critical flaws in life-critical systems may go unnoticed, posing a threat to the safety and well-being of future users.
These actions not only endanger lives but also run contrary to our ethical values. Placing the entire blame on parties that are only partially responsible for an incident contradicts the principles of fairness and accountability. The appropriate distribution of responsibility is crucial for ensuring that the errors are properly addressed and the necessary improvements are made.

\subsection{Implement beliefs}

Explanation providers may use the explanations to promote their belief system, either by influencing people through propaganda or by excluding applicants that they deem unworthy, despite the machine learning model not sharing this perspective.

\subsubsection{Computational propaganda} \label{subsubsec:propaganda}
The power to choose an explanation that best fits its interest, can be used to exert an influence on the public opinion. 
 Propaganda itself is defined as `\emph{the expression of opinion or action by individuals or groups deliberately designed to influence opinion or actions of other individuals or groups with reference to predetermined ends}', while
computational propaganda is defined as `\emph{propaganda created or disseminated using computational (technical) means}'~\citep{martino2020survey}.
Note that propaganda does not necessarily have to lie; it could simply cherry-pick the facts, which is exactly the option explanation providers have to their disposal~\citep{martino2020survey}.
By selectively presenting explanations that align with their preferred ideology or desired narrative, explanation providers can amplify certain perspectives while downplaying or ignoring others.
For example, in the realm of political campaigns, AI systems are used to analyze public sentiment, create targeted messaging, and influence voter behavior. Imagine an entity with access to an AI model that predicts the likelihood of successful integration for immigrants based on various factors like employment, language proficiency, and government support. The entity firmly believes in the principle of stricter requirements for immigrants, and they could selectively highlight specific factors such as language proficiency or employment history, while downplaying or omitting other important factors such as government support and community involvement. By presenting the AI model's predictions as mainly being driven by these selected factors, they could frame the narrative that successful integration is mainly due to language proficiency, and engaging in employment. The goal is to shape public opinion regarding immigration policy and generate support for stricter language and employment requirements for immigrants. Evidently, machine learning models cannot perfectly mimic the actual world, so even if a machine learning model could be perfectly explained, such an explanation would not constitute a perfect explanation of the real world. However, the concern here lies in the fact that people may still perceive machine-generated explanations as accurate depictions of the actual world, and consequently, the cherry-picked explanations have the potential to influence and shape their understanding of the world at large.
Additionally, if the power to generate the explanations would be in the hands of a few actors, they would have the potential to wield significant influence over a large number of people.
In this context, the manipulation of explanations can have far-reaching consequences for public opinion and democratic decision-making. It undermines the principles of transparency and a fair exchange of ideas, and could promote the spread of misinformation.

\subsubsection{Avoid undesired applicants} \label{subsubsec:undesired_app}
In this scenario, the explanation provider, who is using a machine learning model, has the ability to engage in discriminatory practices without directly manipulating the model itself. Instead, they alter the quality of the explanations given to certain population groups, thereby introducing discrimination.
In algorithmic decision-making, explanations are often provided to users (the explanation recipients) to help them understand the factors that influenced the decision and potentially take corrective actions (\emph{algorithmic recourse}). Counterfactual explanations are most often used here, as they guide users in modifying their input data to achieve a desired outcome.

In this case, the explanation provider treats different population groups unequally by manipulating the quality of the explanations provided to them. The preferred population group is given explanations that are concise, actionable, and easily implementable. For example, they might receive suggestions such as adjusting the loan amount slightly or making small changes to their reported income. These explanations empower the preferred group to take specific actions that could potentially improve their chances of receiving a positive outcome.
On the other hand, the disadvantaged demographic group is given explanations of lower quality. These explanations are designed to be difficult or even impossible to act on. They might involve suggesting large changes to their income or modifying their age, which are factors that applicants typically have limited or no control over. By providing such explanations, the explanation provider creates a significant imbalance in the recourse options available to different population groups.
These population groups are not solely confined to traditionally protected characteristics such as race or gender. They can extend to any characteristic that the explanation provider deems undesirable. For example, in the hiring domain, the hiring company (and explanation provider) may deliberately offer lower-quality explanations to older individuals or individuals with certain health conditions, as they perceive them as less desirable for future employment. For some cases, this could also lead to an increase in profit which shows that the multiple objectives can be pursued in parallel and may not always require mutual exclusion.
Note that the discriminatory practices described in this scenario are not related to the machine learning model itself, but to the post-processing stage where explanations are generated and shared with applicants. This issue is related to fairness in algorithmic recourse, where fairness is assessed by measuring the distance between the factual and the counterfactual instance~\citep{von2022fairness,sharma2020certifai}, and highlights the need for fairness assessments not only during the modeling stage but throughout the entire decision-making pipeline, including the provision of explanations.
\vspace{2mm}


\subsection{Increase profit}
Explanation providers might feel incentivized to capitalize on the explanations. They could return the explanation that would be the most profitable for them, and for this we envisage several scenarios.

\subsubsection{Advertising}
One possibility discussed in previous work, is the integration of algorithmic explanations with advertising opportunities, creating an `\emph{explanation platform}' where advertisements are served alongside the explanation~\citep{greene2023monetizing}. 
An example of this could be, that during a job application you receive the following explanation: `\textit{If your CV would have included Python, you would have been invited for the next round}'. This explanation would then be accompanied by an advertisement for an online Python course, which would be a convenient solution for users to reach their goal~\citep{greene2023monetizing}. This approach allows the explanation provider to select the explanations that have the potential to generate the highest revenue in the advertising market. 


\subsubsection{Highlight profit-maximizing explanations}
However, monetization avenues can go beyond advertising. Explanation providers can also exploit the plurality of explanations to direct users towards actions that would maximize their own profits directly. This is related to the advertising scenario, but in this case the actions of the decision subject would directly lead to an increase in profit for the provider.
 For example, in the domain of healthcare diagnostics, AI systems are increasingly used for the identification of diseases and treatment recommendations. Malicious explanation providers, such as healthcare providers or insurance companies, may strategically choose explanations that prioritize certain measures or specific treatments. In this context, the goals of healthcare providers and insurance companies may diverge. Healthcare providers may have incentives to promote more expensive treatments, while insurance companies may prefer cost-saving measures and cheaper treatment options. However, by favoring explanations that are not necessarily the best or most appropriate, these providers can exert influence over medical decisions and potentially compromise patient care. 
 This scenario could also happen in other domains than healthcare: for example, in the realm of credit scoring,  AI systems are employed to evaluate an individual's creditworthiness. \citet{barocas2020hidden} already mention that decisions (and therefore explanations) in this scenario are not simply binary. The provider gives the decision subject a counterfactual that results in a \emph{specific} interest rate, and as such it can choose the interest rate that is likely to maximize its profit~\citep{barocas2020hidden}.   

 \subsubsection{Engage users}
  In line with \emph{Computational Propaganda}, discussed in Section~\ref{subsubsec:propaganda}, providers could also choose to return the explanations that reinforce the ideologies of the data subject itself. In this case, the explanation provider would be a platform, and the goal would be to maximize the revenues of the platform by keeping users as engaged and satisfied as possible (for many platforms daily/monthly active users is an important objective in their financial reports).  An example of an explanation in this case, could be the same as in the scenario of propaganda, but in this case different society groups would receive very different explanations, depending on their beliefs.
 It is known that presenting them with content and information that is likely to resonate with their interests is a way to achieve this (in line with filter bubbles in content recommendation systems). However, this could lead to different groups in society receiving vastly different explanations for the same phenomenon, and consequently to \emph{epistemic fragmentation}~\citep{milano2021epistemic}.\footnote{Epistemic fragmentation refers to the tendency for different people to have different sources of knowledge and different, often conflicting, understandings} 
  By reinforcing filter bubbles and echo chambers, these platforms exacerbate polarization and hinder constructive dialogue between different groups in society.

\vspace{2mm}
 Introducing a profit motive into the generation of explanations at all seems contradictory to the initial goals of Explainable AI. An explanation recipient should not have to wonder whether the selected explanation was chosen for its profit-making potential rather than for its ability to accurately explain the situation~\citep{greene2023monetizing}.

\section{Discussion} \label{sec:discussion}
The examples discussed in Section~\ref{sec:objectives} shed light on potential ethical concerns, even though they may not necessarily involve illegal activities. In these scenarios, the generated explanations remain factually correct but are selectively hand-picked by the explanation provider to serve their own interests. At the moment, this process is completely unregulated, but could have very serious consequences, as outlined in the scenarios above. 

In these scenarios, we assumed the explanation providers had malicious incentives, but obviously, this will not always be the case. In fact, some providers may be motivated to manipulate the explanations for the social good. For example, they might explicitly avoid providing explanations that reinforce biased stereotypes, in an attempt to promote fairness and equity.
Nevertheless, even though their motives might be aligned with societal goals, it remains questionable whether unregulated entities without the required authority should have the power to make this call. 

As we are at the forefront of the XAI revolution, it is crucial to address this issue now, before these methodologies are implemented on an even wider scale.
Currently, a substantial portion of AI power is concentrated among a few tech giants. If we also grant them the authority to control the explanations generated by AI models, they would possess yet another means to exert significant influence over society. To mitigate this concentration and potential misuse of power, it becomes imperative for government institutions to collaborate and establish agreed-upon standards and tools for XAI. In particular, in adversarial contexts where interests may clash, it should not be left solely to the explanation providers to create and choose the explanations. Instead, we argue that governments and policy makers should take the matter into their own hands, and agree on a framework that should be used as soon as possible. The key question here is “\emph{What should be the process to make this decision, and what tools are needed to support this process?}”.  Academic researchers should help in answering this question by proposing a set of tools that can be used,  and by promoting the transparency of digital platforms in their whole process~\citep{greene2022barriers}. 
Similar to the no free lunch theorem, that indicates that there is no algorithm that always outperforms all others, there likely will also not be an universally superior explanation method. An agreement on which method to use in which scenario should be established, and this should be done democratically by allowing those affected by XAI to voice their opinion~\citep{kuzba2020would,vermeire2022choose}, in line with the `democratic principles of affected interests'~\citep{fung2001deepening}. 
 To ensure adherence to ethical values, we also foresee that it would be mandatory to have  external auditors conducting audits of AI systems, explanations, and decision-making processes. These auditors should be independent entities without a vested interest in the outcomes, similar to how audits are conducted in other industries. 

It will take some time to reach a global consensus on the procedures that should be used, and therefore as a short-term solution, regulation should demand full \textbf{transparency} in the used explainability method, and settings. This would not remove all potential for manipulation, but would remove some flexibility for the explanation providers to change this continuously. 
In high-stakes contexts, where transparency is of paramount importance, we argue that the the use of white-box models needs more attention~\citep{goethals2022non}, given the manipulation risks surrounding explanations. 
To conclude, we believe that implementing these measures can ensure that AI systems are developed and deployed in a manner that aligns with  societal values, and foster a more transparent and ethical XAI ecosystem.


\section*{Acknowledgements}
This research was funded by Research Foundation—Flanders grant number 11N7723N.
The authors would like to acknowledge Dieter Brughmans  for providing the example depicted in Table 1, and would like to thank Dieter Brughmans, Ine Weyts, Camille Dams and Travis Greene for their feedback on the paper.

\bibliographystyle{unsrtnat}
\bibliography{references}  






\end{document}